\documentclass[letterpaper]{article} 
\usepackage{aaai2026}  
\usepackage{times}  
\usepackage{helvet}  
\usepackage{courier}  
\usepackage[hyphens]{url}  
\usepackage{graphicx} 
\usepackage{natbib}  
\usepackage{caption} 
\usepackage{algorithm}
\usepackage{algorithmic}
\usepackage{amsfonts}
\usepackage{booktabs}
\usepackage{amsmath}
\usepackage{multirow}
\usepackage{appendix}
\usepackage{cite}

\usepackage{newfloat}
\usepackage{listings}
\DeclareCaptionStyle{ruled}{labelfont=normalfont,labelsep=colon,strut=off} 
\floatstyle{ruled}
\newfloat{listing}{tb}{lst}{}
\floatname{listing}{Listing}
\title{Human-Corrected Labels Learning: Enhancing Labels Quality \\ via Human Correction of VLMs Discrepancies}
\author{
    Zhongnian Li\textsuperscript{1, 2}, 
    Lan Chen\textsuperscript{1}, 
    Yixin Xu\textsuperscript{\rm 1}, 
    Shi Xu\textsuperscript{\rm 1}, 
    Xinzheng Xu\textsuperscript{1, 3}\thanks{Corresponding author.}
}
\affiliations{
    \textsuperscript{\rm 1}School of Computer Science and Technology / School of Artificial Intelligence, China University of Mining and Technology, Xuzhou, China\\
    \textsuperscript{\rm 2}Mine Digitization Engineering Research Center of the Ministry of Education, China University of Mining and Technology, Xuzhou, China\\
    \textsuperscript{\rm 3}State Key Lab. for Novel Software Technology, Nanjing University, Nanjing, China\\

    \{zhongnianli, chenlan, yixinxu, shixu, xxzheng\}@cumt.edu.cn
}

\begin{document}

\maketitle

\begin{abstract}
Vision-Language Models (VLMs), with their powerful content generation capabilities, have been successfully applied to data annotation processes. However, the VLM-generated labels exhibit dual limitations: low quality (i.e., label noise) and absence of error correction mechanisms.
 To enhance label quality, we propose Human-Corrected Labels (HCLs), a novel setting that efficient human correction for VLM-generated noisy labels. As shown in Figure \ref{FIG:1}(b), HCL strategically deploys human correction only for instances with VLM discrepancies, achieving both higher-quality annotations and reduced labor costs. 
 Specifically, we theoretically derive a risk-consistent estimator that incorporates both human-corrected labels and VLM predictions to train classifiers. Besides, we further propose a conditional probability method to estimate the label distribution using a combination of VLM outputs and model predictions. 
 Extensive experiments demonstrate that our approach achieves superior classification performance and is robust to label noise, validating the effectiveness of HCL in practical weak supervision scenarios. 
\end{abstract}

\begin{links}
    \link{Code}{https://github.com/Lilianach24/HCL}
    \link{Extended version}{http://arxiv.org/abs/2511.09063}
\end{links}

\section{Introduction}

\begin{figure*}
    \centering
    \includegraphics[scale=0.5]{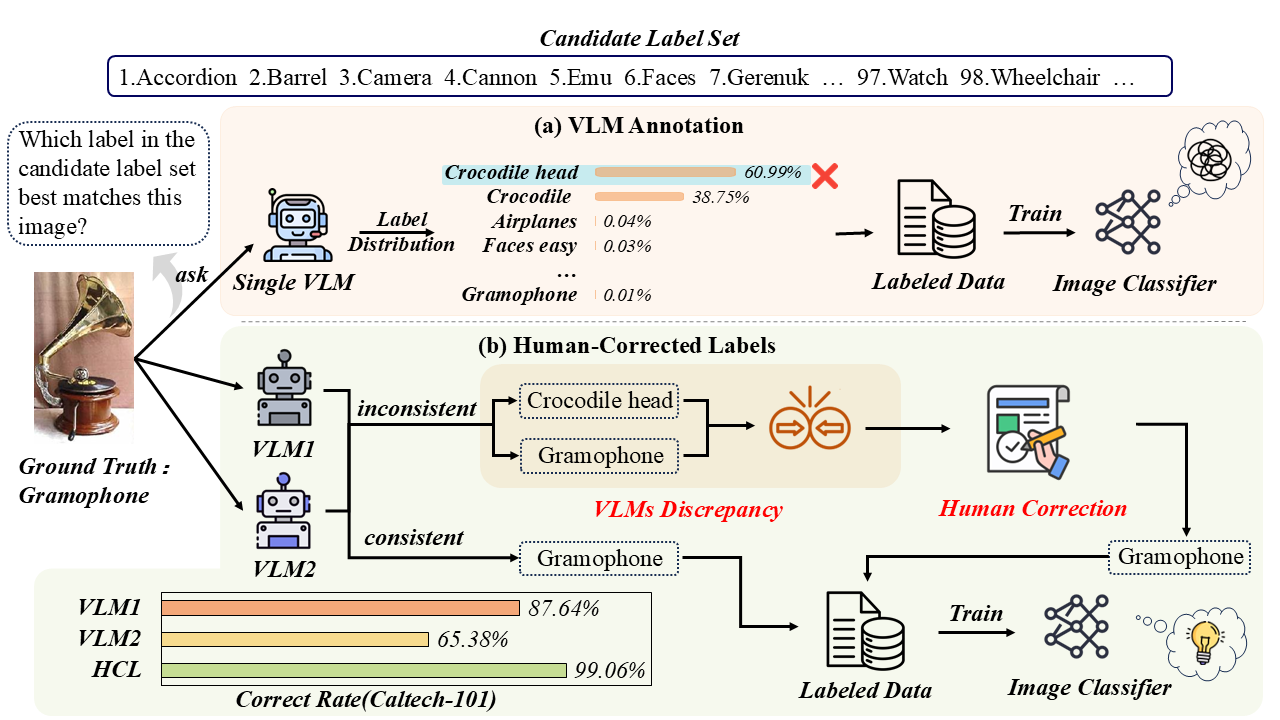}    
    \caption{A comparison between traditional VLM annotation and Human-Corrected Labels (HCLs). The zero-shot results depicted are obtained using CLIP with ViT-L/14. The example images and categories are taken from the Caltech-101 dataset. HCL deploys human correction only for instances with VLMs discrepancies, achieving both higher-quality annotations and reduced labor costs. }    
    \label{FIG:1}
\end{figure*}

Recently, pre-trained VLMs \cite{radford2021learning, li2022blip, liu2023visual} have demonstrated remarkable zero-shot and few-shot recognition capabilities, making them increasingly popular for data annotation tasks. By aligning images and textual labels in a shared embedding space, VLMs can assign labels to images without task-specific training or human correction. This method offers a scalable and cost-effective alternative to traditional annotation procedures, which typically require large amounts of labor and domain expertise. Compared to classical weakly supervised learning techniques \cite{yan2023mutual, chen2023multi, chen2022saliency}, such as semi-supervised learning \cite{van2020survey, cao2021open, wei2024learning}, partial-label learning \cite{feng2020provably, xia2023towards, li2024learning}, and noisy-label learning \cite{menon2015learning, ghosh2017robust, han2020sigua}, VLM-based annotation eliminates the need for initial labeled data, significantly reducing annotation cost and effort \cite{menghini2023enhancing,lilearning}, and enabling rapid dataset construction in new domains.

However, the VLM-generated labels exhibit dual limitations: low quality (i.e., label noise) and absence of error correction mechanisms.
Most notably, existing approaches typically rely on a single VLM to generate labels \cite{kabra2023leveraging}, 
which can lead to biased or incorrect predictions.
As illustrated in Figure~\ref{FIG:1}(a), the VLM may predict “Crocodile head” as the label with the highest confidence, while the ground-truth label is “Gramophone” \cite{wang2022debiased, menghini2023enhancing}. Moreover, current methods lack mechanisms to  correct such errors, making it difficult to ensure label reliability. These limitations highlight the need for new labels strategies that combine the efficiency of VLMs with mechanisms for error detection and correction. 

To address these issues, we propose a setting called Human-Corrected Labels (HCLs).  HCLs strategically deploys human correction only for instances with VLMs discrepancies, thereby achieving both higher-quality annotations and reduced labor costs. Specifically, during annotation, our HCL method automatically accepts labels when multiple VLMs yield consistent predictions, invokes human correction solely for samples where their predictions are inconsistent. As illustrated in Figure~\ref{FIG:1} (b), when annotating an image of a “Gramophone”, correction is only required if there exists a discrepancy among the VLMs. This targeted human correction drastically reduces annotation costs while ensuring high annotation quality. To the best of our knowledge, HCL is the first setting that leverages VLMs discrepancies to correct labels.

In this paper, we develop a risk-consistent learning method that unifies VLM predictions and human-corrected labels in a principled way. Specifically, we derive a risk-consistent estimator that incorporates both human corrected labels and VLMs predictions, enabling effective learning under label uncertainty. Furthermore, we introduce a conditional probability formulation to estimate the label distribution using a combination of VLM outputs and model predictions, enhancing label robustness under weak supervision. Extensive experiments on various datasets clearly demonstrate the effectiveness of the proposed HCL learning method.

Our main contributions are summarized as follows:

\begin{itemize}
\item We propose a novel weakly supervised annotation setting, i.e., Human-Corrected Labels, which combines multiple VLMs with selective human correction. This setting enables high-quality, scalable annotation with minimal human effort.
\item We propose a risk-consistent method that models the conditional label distribution under the HCLs setting, leveraging both  VLM predictions and human corrections.
\item Extensive experiments on standard benchmarks demonstrate that our method significantly improves classification performance and robustness compared to existing weakly supervised baselines.
\end{itemize}

\section{Related Work}
\subsection{Weakly Supervised Learning} 
Weakly supervised learning aims to train predictive models using data with limited or imperfect annotations, and encompasses three main methods: semi-supervised learning \cite{chen2023softmatch}, partial-label learning \cite{chapel2020partial, wang2022freematch}, and noisy label learning \cite{xia2022extended, gong2022class}.
Semi-supervised learning leverages a small amount of labeled data and abundant unlabeled data to improve performance. Key strategies include entropy minimization \cite{grandvalet2004semi, lee2013pseudo}, consistency regularization \cite{sajjadi2016regularization, tarvainen2017mean, miyato2018virtual}, and holistic training methods \cite{berthelot2019mixmatch, sohn2020fixmatch, sosea2023marginmatch}.
Partial-label learning assumes each instance is associated with a set of candidate labels, only one of which is correct. Two main strategies are commonly used: average-based disambiguation \cite{zhang2015solving}, and identification-based approaches that treat the ground-truth as a latent variable \cite{wang2019adaptive, feng2019partial}.
Noisy label learning addresses the challenge of corrupted supervision. Techniques include sample selection \cite{patel2023adaptive}, label correction \cite{albert2023your}, and noise-aware regularization \cite{liu2020early, sun2021co, li2021learning}.

While prior work has advanced weakly supervised learning, it typically assumes fixed supervision and overlooks the potential of combining VLMs with human correction. In contrast, we propose the Human-Corrected Label method, which can achieve the collaboration between VLMs and humans.


\begin{table*}[t]
    \centering
    \begin{tabular}{cccccccc}
        \toprule
        Dataset & Train Samples & Classes & CLIP & Qwen & VLM-CR & CCP & HCLs \\
        \midrule
        CIFAR100       & 50,000  & 100 & 75.66 & 45.82 & 44.02 & 92.97 & 96.90 \\
        Tiny-ImageNet  & 100,000 & 200 & 72.95 & 38.72 & 35.88 & 94.59 & 98.06 \\
        Caltech-101    & 6,074   & 102 & 87.64 & 65.38 & 64.36 & 98.54 & 99.06 \\
        Food-101       & 75,750  & 101 & 89.77 & 48.87 & 49.26 & 96.48 & 98.27 \\
        EuroSAT        & 18,900  & 10  & 43.21 & 39.42 & 26.43 & 81.82 & 95.20 \\
        DTD            & 2,800   & 47  & 52.27 & 56.65 & 48.04 & 81.33 & 91.03 \\
        \bottomrule
    \end{tabular}
    \caption{Annotation accuracy and consistency analysis under the HCLs setting across six benchmarks. Columns show number of training samples and classes, accuracy of CLIP and Qwen labels, VLMs Consistency Rate (VLM CR), Correctness in Consistent Predictions (CCP), and final annotation accuracy after discrepancy-based human correction.}
    \label{tab:1}
\end{table*}

\subsection{Model Discrepancies} 
The model discrepancy, defined as the divergence between predictions of different models, has been widely used in weakly supervised learning to estimate label uncertainty and guide pseudo‑label handling. 
Early disagreement-based methods such as co‑training \cite{wang2017theoretical} and tri‑training \cite{huang2021novel} leverage multiple classifiers (i.e., often on different feature views or model initializations) to label unlabeled data based on inter-model agreement. 
More recent methods, such as Co‑Teaching+ \cite{yu2019does}, combine model discrepancies with small-loss selection to filter noisy supervision. Dynamic Mutual Training \cite{feng2022dmt} further exploits discrepancy for dynamic loss re-weighting, reducing the impact of erroneous VLM-generated labels. 
In semi-supervised object detection, CrossRectify \cite{ma2023crossrectify} uses discrepancies across two detectors to cross-correct VLM-generated labels and improve detection quality. 

However, these methods primarily leverage model discrepancy to improve training robustness, rather than to guide data annotation. Consequently, we utilize discrepancies between multiple pre-trained VLMs (e.g., CLIP \cite{radford2021learning}, Qwen \cite{Qwen2VL}) to identify inconsistent predictions and selectively invoke human correction.

\section{Methods}
In this section, we introduce a detailed description of Human-Corrected Labels. In addition, we propose a risk-consistent estimator to explore and leverage the underlying correlations between the probability distribution of VLM-generated labels and HCLs. 
\subsection{Human-Corrected Labels} 

The predictions produced by VLMs exhibit dual limitations: low quality (i.e., label noise) and the absence of error correction mechanisms. Specifically, VLM predictions may contain incorrect or ambiguous labels,  leading to noisy supervision that can degrade the performance of models trained under weak supervision. As shown in Figure ~\ref{FIG:1}(a), VLMs may annotate the instance as “crocodile head” with the highest confidence under a zero-shot setting, which can result in noisy labels.

In this paper, we propose a new setting: Human-Corrected Labels (HCLs), where labels are generated by VLMs, and annotators only correct instances with inconsistent VLM predictions. For example, as illustrated in Figure ~\ref{FIG:1}(b), when the candidate label set includes \{“accordion”, “barrel”, “camera”, “cannon”, “emu”, “faces”, “gerenuk”, ... “watch”, “wheelchair”, ...\}, for an image of a “gramophone”, the annotator only needs to provide the correct label if the VLMs produce inconsistent predictions; otherwise, the predictions are accepted directly.  Compared to traditional manual annotation, HCL leverages VLM knowledge for high-quality labels while cutting annotation costs.


\subsubsection{Effectiveness of HCLs.} To evaluate the effectiveness of HCLs in improving label quality, we annotate six standard classification benchmarks using CLIP (ViT-L/14) \cite{radford2021learning} and Qwen2.5-VL (7B-Instruct) \cite{Qwen2VL}, and report results in Table~\ref{tab:1}. We compute the consistency rate between VLMs and the correctness rate within the consistent predictions. 
The results indicate that although the agreement between the two VLMs varies across datasets, the labels on which they agree are highly reliable, demonstrating that model consensus is a strong indicator of correctness.
Based on this, HCLs performs human correction only on samples with VLM discrepancies. As shown in the “HCLs” column, the resulting labels achieve consistently high accuracy, validating that targeted human correction guided by model discrepancy yields reliable supervision with minimal human effort.

\subsection{Problem Formalization}  
\subsubsection{Ordinary Labels.} In multi-class classification, let $\mathcal{X} \in \mathbb{R}^d$ denote the feature space and $\mathcal{Y} = \{1, \ldots, k\}$ denote label space, where $d$ is the feature dimension and $k > 2$ is the number of classes. We assume that $\{(x_l, y_l)\}_{l=1}^{N}$ are sampled independently from an unknown distribution with density $p(x, y)$. The goal of ordinary multi-class classification is to learn a classifier $f (x) : \mathcal{X} \rightarrow \mathcal{Y}$ that minimizes the classification risk under a multi-class loss $\mathcal{L}(f (x), y)$ :
\begin{equation}
\begin{aligned}
\label{eq:1}
R(f) 
     &= \mathbb{E}_{x \sim p(x)} \sum_{i=1}^{k} P(y=i \mid x) \mathcal{L}[f(x), i] \\
\end{aligned}
\end{equation}
where $x$ denotes an input image, $p(x)$ denotes the density of $x$,  $y$ denotes its ground truth label, and $P(y=i \mid x)$ denotes the probability that the true label of $x$ is class $i$.

\subsubsection{HCLs.}

To address the limitations of traditional VLM-generated labels, we propose using multiple VLMs to collaboratively annotate data, leveraging their complementary strengths while detecting potential label noise through discrepancy signals. 
Given a training set with HCL annotations, denoted as $\mathcal{D}_{HCL} = \{(x_j, Y_j, s_j)\}_{j=1}^{N}$, where $x_j$ is the input, and $s_j \in \{0,1\}$ indicates whether the VLM predictions are consistent (i.e., $s_j = 0$) or inconsistent (i.e., $s_j = 1$). The HCL label $Y_j$ is defined as:
\begin{equation}
\label{eq:2}
Y_j = 
\left\{
\begin{array}{ll}
\hat{y}_j \,\, \text{or} \,\, \tilde{y}_j, & \hat{y}_j = \tilde{y}_j \text{ and } s_j = 0 \\
y_j, & \hat{y}_j \neq \tilde{y}_j  \text{ and } s_j = 1  \\ 
\end{array}
\right. ,
\end{equation}
where $\hat{y}_j$ and $\tilde{y}_j$ represent the predicted labels from two different VLMs, respectively.

Our goal is to develop a classifier $f(x)$ that effectively utilizes the complementary pseudo-label signals from multiple VLMs while leveraging the discrepancy indicator to mitigate the impact of noisy labels, thus enhancing the reliability and scalability of VLM-based supervision for multi-class image classification.

\subsection{Risk-Consistent Estimator}
Based on the proposed HCL setup, we present a risk-consistent learning method \cite{feng2020provably, feng2020learning, xu2022one}. To rigorously connect the ground-truth label and HCL labels, we introduce the following definition.

\subsubsection{Definition 1. (HCL Condition.)} Given a  sample $(x, Y, s) \in \mathcal{D}_{\mathrm{HCL}}$, $\forall i,j \in \mathcal{Y}$, we define the conditional probability of HCLs to satisfy:
\begin{equation}
\begin{split}
\label{as:1}
 P(y = i, Y = j \neq i, s = 1 | x)  &= 0.  \\
\end{split}
\end{equation}
This definition implies that for samples with VLM discrepancies ($\hat{y}_i \neq \tilde{y}_i$, i.e., $s_j=1$), the ground-truth label $y$ must be different from the label predicted by at least one VLM ( i.e., $\hat{y}$ or $\tilde{y}$).
For these instances, we rely on human correction to provide the true label $y$ as a trustworthy supervision label. 
Although the true conditional distribution $P(y = i \mid x)$ in Eq.~\ref{eq:1} is unknown, we approximate it under the HCL probabilities  on consistent predictions (i.e., $s = 0$) as:
\begin{equation}
\label{as:2}
\{ P(y = i, Y, s = 0 \mid x) \}_{i=1}^{k}, \quad m = 1, \ldots, k.
\end{equation}

\noindent \textbf{Lemma 2.} Under the HCL Definition 1, the conditional probabilities $P(y=i \mid x)$ can be expressed as:
\begin{equation}
\label{le:2}
\begin{aligned}
P(y=i \mid x) &= P(Y = i, s = 1 \mid x) \\
        & \quad + \sum_{m=1}^{k} \Big[ P(y = i\mid Y = m, s = 0,  x) \\
        & \qquad  P (Y=m, s=0 \mid x) \Big].\\
\end{aligned}
\end{equation}
This decomposition expresses the total probability of class $i$ given instance $x$ as the sum of two components: the probability of human correction, and a weighted conditional probability by the marginal probability. This formulation enables the model to express the unknown true label distribution using a combination of corrected labels from inconsistent samples and soft VLM-generated labels from consistent samples. The proof is provided in the Appendix A.1.

\noindent \textbf{Theorem 3.}
To address the HCL learning problem, according to the Definition 1 and Lemma 2, the classification risk $R(f)$ in Eq.~\ref{eq:1} can be reformulated as:
\begin{equation}
\label{eq:3}
\begin{aligned}
R_{HCL}(f) &=  \mathbb{E}_{(x, Y, s=1)} \mathcal{L}[f(x), i] + \\
& \mathbb{E}_{(x, Y, s=0)} \sum_{i=1}^{k} P(y=i \mid Y, s=0, x) \mathcal{L}[f(x), i],
\end{aligned}
\end{equation}
where $P(y = i \mid Y, s = 0, x)$ denotes the conditional probability derived from consistent VLM predictions, enabling the classifier $f$ to learn from a conditional label distribution. In contrast, for samples with discrepancies (i.e., $s = 1$), the classifier is trained using human-corrected labels to ensure reliable supervision. The proof is provided in the Appendix A.2.

\subsection{Practical Implementation}
\subsubsection{Loss Function.} 
To effectively leverage VLM-generated labels while mitigating noise under weak supervision, we design a tailored loss function aligned with our VLM-consistency filtering setting. Specifically, we adopt a modified MSE-like loss, formulated explicitly to enhance discriminability and training stability. Let the classifier output be $f(x) \in \mathbb{R}^k$. 
We use the loss as:
\begin{equation}
\mathcal{L}(f(x), i) = \frac{1}{k} \sum_{j=1}^{k} \left[1 - f_j(x)  (2 \delta_{ij} - 1) \right]^2 ,
\end{equation}
where $f_j(x)$ denotes the $j$-th classifier, and $\delta_{ij}$ is the Kronecker delta (i.e., $\delta_{ij} = 1$ if $i = j$, and $\delta_{ij} = 0$, otherwise). Specifically, the correct class (where $i = j$) is assigned $1$ to emphasize its importance, while incorrect classes (where $i \neq j$) are assigned $-1$ to suppress their activation. The use of squared error penalizes both under-activation for the correct class and over-activation for incorrect classes, encouraging sharper class separation. 

\subsubsection{Empirical Approximation.} We arrange the HCLs dataset as $\mathcal{D}_c = \mathcal{D}_H \cup \mathcal{D}_V$, where $\mathcal{D}_H$ and $\mathcal{D}_V$ denote the set of VLM-inconsistent data and VLM-consistent data respectively. Then, the empirical approximation of the classification risk in Eq.~\ref{eq:3} is given by:
\begin{equation}
\begin{aligned}
\hat{R}_{\text{HCL}}(f) &= \frac{1}{|\mathcal{D}_H|} \sum_{x_j \in \mathcal{D}_H} \mathcal{L}(f(x_j), y_j) \\
                        & + \frac{1}{|\mathcal{D}_V|} \sum_{x_j \in \mathcal{D}_V} \sum_{i=1}^{k} \Big[ \hat{P}(y_j=i \mid Y_j, s_j=0, x_j) \\
                        & \qquad \mathcal{L}(f(x_j), i) \Big] ,
\end{aligned}
\end{equation}
where $\hat{P}(y_j=i \mid Y_j, s_j=0, x_j)$ is the estimated conditional label distribution for sample $x_j$.


This empirical approximation enables effective learning under weak supervision while controlling  VLM-generated label noise, thereby supporting scalable training on large-scale unlabeled datasets with minimal human correction.

\begin{table*}[ht]
    \centering
    \begin{tabular}{ccccccc}
        \toprule
                & CIFAR100 & Tiny-ImageNet & Caltech-101 & Food-101 & EuroSAT & DTD \\
        \midrule
        & \multicolumn{6}{c}{Supervised Labels Learning} \\
        \cmidrule(lr){2-7}
        FSL-C \cite{radford2021learning} & 83.12 & 81.13 & 95.77 & 93.14 & 95.96 & 76.02 \\
        \cmidrule(lr){1-7}
        & \multicolumn{6}{c}{Weakly Supervised Learning Methods} \\
        \cmidrule(lr){2-7}
        Zero-shot CLIP \cite{radford2021learning} & 73.90 & 72.29 & 88.45 & 88.21 & 47.35 & 51.98 \\
        HL-C \cite{radford2021learning} & 74.04 & 72.50 & 57.54 & 73.90 & 95.48 & 60.90 \\
        VL-C \cite{radford2021learning} & 66.25 & - & 42.02 & 73.66 & - & 45.54 \\
        Only-C \cite{radford2021learning} & 77.11 & - & 50.44 & 92.48 & - & 52.22 \\
        Only-Q \cite{Qwen2VL} & 49.77 & - & 35.53 & 54.90 & 11.64 & 55.35 \\
        DIRK \cite{wu2024distilling} & 77.26 & 52.91 & 63.66 & 90.12 & 44.67 & 67.46 \\
        PaPi \cite{xia2023towards} & 79.67 & 25.33 & 57.66 & 90.73 & 29.01 & 63.62 \\
        HCL (Ours) & \textbf{82.48} & \textbf{79.87} & \textbf{94.68} & \textbf{93.07} & \textbf{95.53} & \textbf{70.76} \\
        \bottomrule
    \end{tabular}
    \caption{Accuracy (\%) of all compared methods on six datasets. The highest value for each dataset is highlighted in bold. Results below 10\% are omitted.}
    \label{tab:2}
\end{table*}

\subsubsection{Conditional Probability Estimation.}
Minimizing the empirical risk $\hat{R}_{HCL}$, as introduced in Theorem 3, requires estimating the conditional probability distribution $P(y=i \mid Y, s=0, x)$ for VLM-consistent samples. To leverage both the prior knowledge from VLMs and the model predictions, we estimate this distribution via a weighted distribution of two sources. The model-based conditional distribution $P_{model}$ is obtained by applying the softmax function to the classifier output logits $f(x) \in \mathbb{R}^k$:
\begin{equation}
\begin{aligned}
    P_{\text{model}}(y=i \mid x) = \frac{e^{f_i(x)}}{\sum_{l=1}^{k} e^{f_l(x)}}.
\end{aligned}
\end{equation}
The external distribution $P_{CLIP}$ is derived from the CLIP model via image-text alignment. Specifically, for an image $x$, we compute its image embedding $g_I(x)$ and compare it with a set of pre-encoded text embeddings $Q_T = [q_1, \dots, q_k]$, where each $q_i$ corresponds to class $i$. The resulting probability distribution is computed using cosine similarity followed by temperature-scaled softmax:
\begin{equation}
\begin{aligned}
P_{\text{CLIP}}(y=i \mid x) = \frac{e^{\tau \cdot \cos(g_I(x), q_i)}}{\sum_{l=1}^{k} e^{\tau \cdot \cos(g_I(x), q_l)}}.
\end{aligned}
\end{equation}
where $\cos(\cdot, \cdot)$ denotes cosine similarity and $\tau > 0$ is a temperature parameter used to adjust confidence sharpness. Empirically, we set $\tau = 100$ to produce confident yet stable predictions.

To form the final estimated conditional distribution $\hat{P}(y=i \mid Y, s=0, x)$, we compute a convex combination of $P_{CLIP}$ and $P_{model}$, weighted by a mixing coefficient $\lambda \in [0,1]$:
\begin{equation}
\begin{aligned}
\hat{P}(y=i \mid Y, s=0, x) = \lambda P_{CLIP} + (1 - \lambda) P_{model}.
\end{aligned}
\end{equation}
This interpolation enables dynamic balancing between external VLM-generated labels and the evolving predictions of the model, which is particularly beneficial under weak supervision.

\subsubsection{Model.} 
We employ the CLIP model with ViT-L/14 \cite{radford2021learning} as the vision backbone for feature extraction, followed by a trainable linear classification head. Specifically, we utilize the open-sourced CLIP implementation in FP32 precision, where the image encoder is frozen to act as a feature extractor during the entire training process. The extracted visual features are of dimension 768, which are then fed into a learnable linear layer for classification. This design allows us to leverage the strong representation capabilities of pre-trained VLMs while maintaining computational efficiency.

\section{Experiments}
\subsection{Experimental Setup}
\subsubsection{Datasets.} To comprehensively evaluate the effectiveness of our proposed method, we conduct experiments on six diverse multi-class image classification datasets, spanning both coarse-grained (i.e., CIFAR-100 \cite{krizhevsky2009learning}, Tiny-ImageNet \cite{le2015tiny}, and Caltech-101 \cite{fei2004learning}) and fine-grained (i.e., Food-101 \cite{bossard2014food}, EuroSAT \cite{helber2019eurosat}, and DTD \cite{cimpoi2014describing}) classification tasks across different domains. All images are resized to $224 \times 224$ before being fed into the models. Detailed descriptions of the datasets are provided in the Appendix A.3.

\begin{figure*}
    \centering
    \includegraphics[scale=0.5]{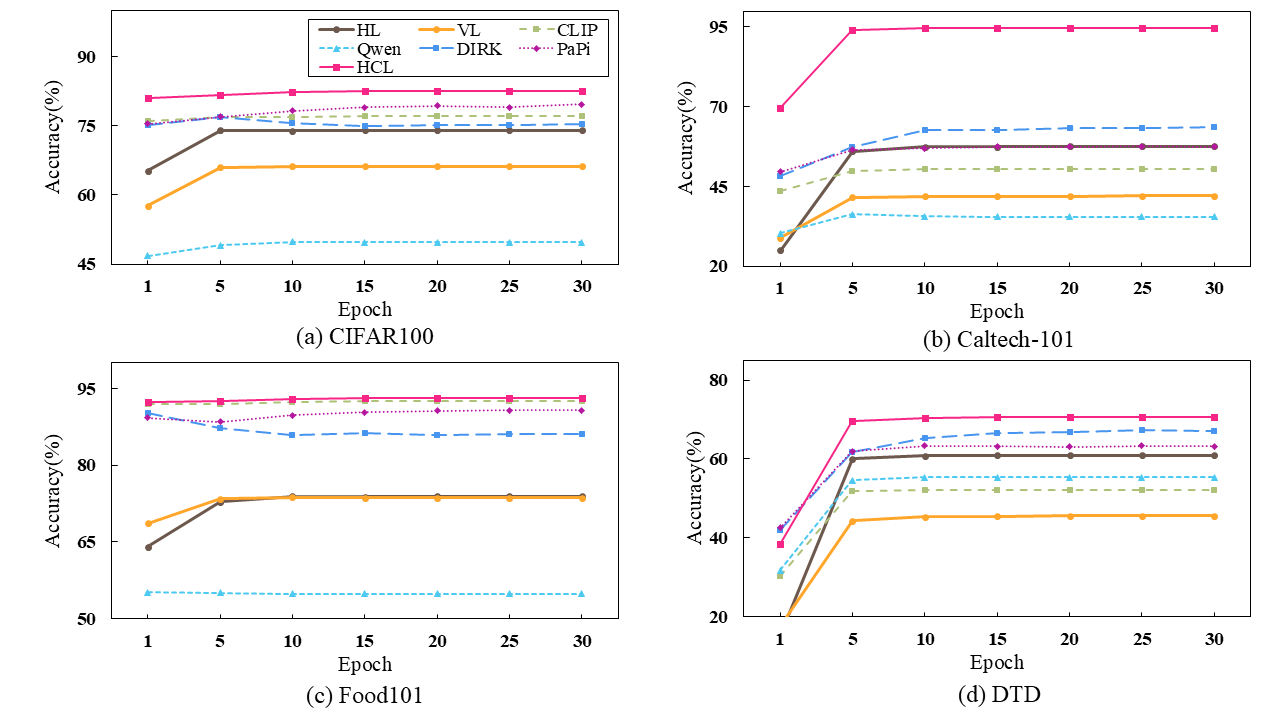}    
    \caption{Classification accuracy over training epochs (1–30) for each method on (a) CIFAR100, (b) Caltech-101, (c) Food-101, and (d) DTD.}
    \label{FIG:2}
\end{figure*}

\subsubsection{Compared Methods.} 
We evaluate the proposed HCL method against representative baselines, including state-of-the-art weakly supervised methods and VLM-based linear probing under varying supervision settings:
\begin{itemize}
    \item DIRK \cite{wu2024distilling} and PaPi \cite{xia2023towards}: Partial-label learning methods using VLM predictions to construct candidate label sets. For consistent samples (i.e., $s_j=0$), only the CLIP label is used; for inconsistent ones (i.e., $s_j=1$), the set includes predictions from CLIP, Qwen, and the human-corrected label. 
    \item FSL-C \cite{radford2021learning}: Fully supervised method trained on all labeled training data using ground-truth.
    \item HL-C \cite{radford2021learning}: Training only on samples with VLM discrepancies (i.e., $s_j=1$), using human-corrected labels; consistent samples are discarded.
    \item VL-C \cite{radford2021learning}: Weak supervision method using VLM predictions on consistent samples only (i.e., $s_j=0$).
    \item Only-C \cite{radford2021learning} and Only-Q \cite{Qwen2VL}: Training solely with CLIP or Qwen predictions, without considering consistency.
\end{itemize}

These baselines enable a thorough comparison under unified settings, highlighting the effectiveness of our hybrid supervision approach guided by VLM discrepancies.
\subsubsection{Implementation Details.} 
All experiments are implemented in PyTorch and conducted on a single NVIDIA RTX 4090 GPU. We fix the random seed to 42 to ensure reproducibility. All models are trained for 30 epochs with a batch size of 64 using the AdamW optimizer \cite{loshchilov2017decoupled} with a learning rate of $5 \times 10^{-4}$ and a weight decay of $1 \times 10^{-4}$. To ensure stable convergence, we adopt a step-wise learning rate scheduler that decays the learning rate by a factor of 0.1 every 5 epochs. This lightweight and consistent training setup ensures scalability and efficiency across all datasets and methods. Unless otherwise specified, the hyperparameter $\lambda$ is typically set to 1.0 by default.

\subsection{Comparison with State-of-the-Art}
To evaluate the effectiveness of the proposed HCL method, we conduct experiments on six benchmark datasets under weakly supervised methods. Label distributions are generated by CLIP with ViT-L/14 and used within HCL to train a linear classifier with minimal human correction. All experiments are trained with identical configurations for fair comparison, and results are summarized in Table \ref{tab:2}.

For clarity, results with accuracy below 10\% are omitted. In these cases, methods relying solely on VLM-generated labels (i.e., VL-C, Only-C, Only-Q) perform poorly on challenging datasets such as Tiny-ImageNet and EuroSAT. This indicates that raw VLM predictions, without human correction or filtering, are unreliable under domain shifts.

Across all datasets, HCL consistently outperforms existing weakly supervised baselines. On challenging datasets like Tiny-ImageNet and EuroSAT, it achieves substantial improvements over DIRK and PaPi, illustrating the effectiveness of combining VLM priors with targeted human correction. Moreover, HCL approaches the performance of fully supervised models, narrowing the gap between weak and full supervision while greatly reducing annotation cost.

In summary, HCL leverages VLM priors and selectively applies human correction, providing reliable supervision at low cost. By using label distributions as richer training signals, HCL captures finer category boundaries, making it a scalable and practical approach for real-world weakly supervised image classification.

\subsection{Comparison of Training Cost}
To assess the efficiency of HCL, we compare training convergence across four datasets: CIFAR100, Caltech-101, Food-101, and DTD. Figure~\ref{FIG:2} shows accuracy versus training epochs for HCL, CLIP linear probing, Only-Q probing, DIRK, and PaPi.

HCL achieves faster convergence and higher final accuracy than all baselines. It stabilizes within 10 epochs, whereas methods such as DIRK and PaPi require more iterations and converge to lower performance. These results indicate that HCL enhances both training efficiency and accuracy, providing a computationally efficient solution for large-scale weakly supervised learning.

\begin{figure}
    \centering
    \includegraphics[scale=0.4]{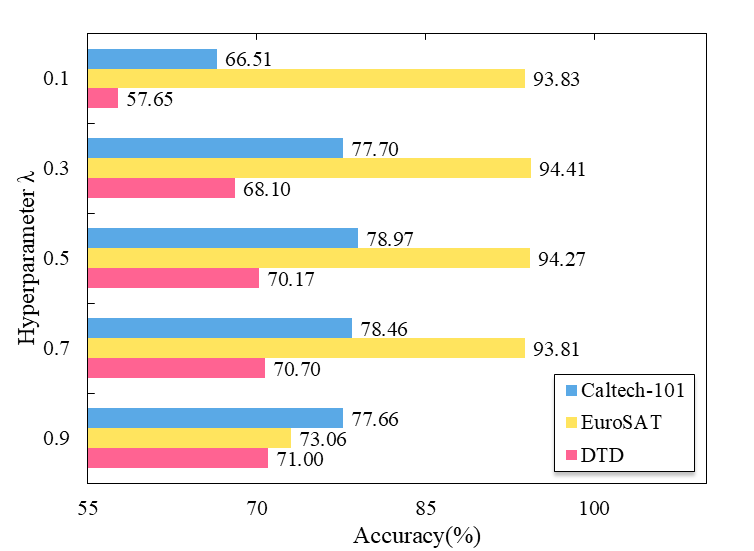}    
    \caption{Experimental results on the influence of the conditional probability hyperparameter $\lambda$. Experiments are performed on human corrected HCLs data.}
    \label{FIG:3}
\end{figure}

\subsection{Influence of The Hyperparameter $\lambda$}
We examine how performance varies with $\lambda$ on Caltech-101, EuroSAT and DTD. Figure~\ref{FIG:3} shows that as $\lambda$ increases, introducing conditional probability distribution does not improve HCL performance on Caltech-101 and EuroSAT; instead, accuracy declines. This is expected, as CLIP already provides highly reliable supervision here, and additional self-predictions may introduce noise. However, on DTD, distribution yields slight but consistent improvements over standard HCL. These results indicate that on domain-shifted or texture-centric datasets, where CLIP supervision may have domain bias or uncertainty, incorporating adaptive model predictions refines and corrects noisy or biased VLM-generated labels, enhancing supervision quality.

\subsection{Impact of Various VLMs}

To investigate how the choice and number of VLMs affect the consistency-based filtering in HCL, we conduct research by varying the VLM combinations used to assess prediction agreement. Specifically, we consider three configurations: (1) Case 1, the default setup using CLIP and Qwen, (2) Case 2, an alternative using CLIP and LLaVA \cite{liu2023visual}, and (3) Case 3, an extended configuration incorporating CLIP, Qwen, and LLaVA. In two-VLM settings, we report the proportion of consistent predictions; in the three-VLM setting, we report the proportion of fully inconsistent cases (i.e., $s_j = 1$), where all models disagree. For $s_j = 0$, we adopt the majority-agreed label as supervision.

\begin{table}[ht]
    \centering
    \begin{tabular}{lccc}
        \toprule
          & CIFAR100 & Caltech-101 & DTD \\
        \midrule
        Case 1 & 82.48/44.02 & 94.68/64.36 & 70.76/48.04 \\
        Case 2 & 82.42/46.68 & 93.33/38.39 & 69.46/33.19 \\
        Case 3 & 82.42/33.02 & 74.16/28.73 & 70.70/36.61 \\
        \bottomrule
    \end{tabular}
    \caption{HCLs classification accuracy (left) and consistent probability or full-model discrepancy rate (right) under different VLM configurations.}
    \label{tab:4}
\end{table}

\begin{table}[ht]
    \centering
    \fontsize{9}{1em}\selectfont
	\tabcolsep=0.5em
    \begin{tabular}{lccc}
        \toprule
         & CIFAR100/VLM & Caltech-101/VLM & DTD/VLM \\
        \midrule
        Prompt A & 82.48/45.82 & 94.68/65.38 & 70.76/56.65 \\
        Prompt B & 82.48/44.69 & 94.83/43.00 & 71.06/57.66 \\
        Prompt C & 82.40/43.57 & 94.24/38.91 & 71.35/60.86 \\
        \bottomrule
    \end{tabular}
    \caption{HCLs classification accuracy (left) and corresponding VLM-generated label accuracy (right) under different prompts across datasets.}
    \label{tab:5}
\end{table}

As shown in Table~\ref{tab:4}, accuracy remains stable across CIFAR100 and DTD, indicating that HCLs is generally robust to the choice of VLMs. However, on Caltech-101, performance declines under Case 3, likely due to the relatively low proportion of fully inconsistent samples, meaning many training samples rely on partially consistent predictions from two VLMs. If the VLMs themselves make incorrect predictions, this increases the likelihood of introducing noisy labels. This suggests that while HCLs supports flexible VLM integration, more models do not always yield better results.

\subsection{Impact of Various Prompts}
To evaluate the sensitivity of the HCLs setting to prompt design when generating VLM-generated labels, we study the use of different prompt variants. Specifically, we compare three types: (1) Prompt A and Prompt B, same content in different languages, and (2) Prompt C, a simplified English version with concise phrasing. Detailed prompt descriptions are shown in Appendix A.4.

As shown in Table~\ref{tab:5}, prompt choice affects pseudo-label accuracy, but HCLs maintains stable final performance across datasets.
This robustness stems from its use of model discrepancies and human correction, allowing tolerance to prompt-induced noise. These results suggest HCLs reduces the need for extensive prompt tuning when using general-purpose VLMs.

\section{Conclusion}
In this paper, we investigate a novel weakly supervised learning setting, Human-Corrected Labels (HCLs), which strategically incorporates minimal human corrections to improve the quality of noisy VLM-generated labels. We formulate a risk-consistent learning method that leverages both VLM predictions and human corrections by estimating conditional label distributions, enabling effective learning under label noise. Extensive experiments across standard and real-world datasets demonstrate that our method achieves high classification accuracy while substantially reducing annotation cost, highlighting the practicality and robustness of the proposed HCLs.

\section{Acknowledgments}
This work was supported by the National Natural Science Foundation of China (No.62306320, 61976217), the Open Project Program of State Key Lab. for Novel Software Technology (No. KFKT2024B32), and the Natural Science Foundation of Jiangsu Province (No. BK20231063).


 \onecolumn
 \appendix
 \section{A. Appendix / supplemental material}
 \subsection{A.1 Proof of Lemma 2}
 \label{proof lemma}
 \noindent \textbf{Lemma A.1.} Under the HCL Definition 1, the conditional probabilities $P(y=i \mid x)$ can be expressed as:
 \begin{equation}
 \begin{aligned}
 P(y=i \mid x) &= P(Y = i, s = 1 \mid x) + \sum_{m=1}^{k} \Big[ P(y = i\mid Y = m, s = 0,  x)  P (Y=m, s=0 \mid x) \Big].\\
 \end{aligned}
 \end{equation}
 \textit{Proof.} By the law of total probability:
 \begin{equation}
 \begin{aligned}
 P(y=i \mid x) &= P(j = i, s = 1 | x) + P(j = i, s = 0 | x) \\
         &= \sum_{m = 1}^k P(y=i, Y = m, s = 1 \mid x) + \sum_{m=1}^k P(y=i, Y=m, s=0) \\
         &= P(y=i, Y=i, s=1 \mid x) + \sum_{m=1}^{k} \Big[ P(y = i\mid Y = m, s = 0,  x)  P (Y=m, s=0 \mid x) \Big] \\
         &= P(Y = i, s = 1 \mid x) + \sum_{m=1}^{k} \Big[ P(y = i\mid Y = m, s = 0,  x)  P (Y=m, s=0 \mid x) \Big]\\
 \end{aligned}
 \end{equation}

 \subsection{A.2 Proof of Theorem 3}
 \label{proof theorem}
 \noindent \textbf{Theorem A.2.}
 To address the HCL learning problem, according to the Definition 1 and Lemma 2, the classification risk $R(f)$ in Eq.1 can be reformulated as:
 \begin{equation}
 \begin{aligned}
 R_{HCL}(f) &=  \mathbb{E}_{(x, Y, s=1)} \mathcal{L}[f(x), i] + \mathbb{E}_{(x, Y, s=0)} \sum_{i=1}^{k} P(y=i \mid Y, s=0, x) \mathcal{L}[f(x), i].
 \end{aligned}
 \end{equation}
 \textit{Proof.} According to Definition 1 and Lemma 2
 \begin{equation}
 \begin{aligned}
 R_{HCL}(f) &=  \mathbb{E}_{x \sim p(x)} \sum_{i=1}^{k} P(y=i \mid x) \mathcal{L}[f(x), i] \\
             &= \mathbb{E}_{x \sim p(x)} \sum_{i=1}^k \sum_{j=0}^1 P(y=i, s=j|x) \mathcal{L}[f(x), i] \\
             &= \mathbb{E}_{x \sim p(x)} \sum_{i=1}^k P(y=i, s=1 \mid x) \mathcal{L}[f(x), i] + \mathbb{E}_{x \sim p(x)} \sum_{i=1}^k P(y=i, s=0 \mid x) \mathcal{L}[f(x), i] \\
             &= \mathbb{E}_{x \sim p(x)} \sum_{i=1}^k \sum_{m = 1}^k P(y=i, Y = m, s = 1 \mid x) \mathcal{L}[f(x), i] + \mathbb{E}_{x \sim p(x)} \sum_{i=1}^k \sum_{m=1}^k P(y=i, Y=m, s=0 \mid x) \mathcal{L}[f(x), i] \\
             &= \mathbb{E}_{x \sim p(x)} \sum_{i=1}^k P(y=i, Y=i, s=1 \mid x) \mathcal{L}[f(x), i] \\
             & \quad + \mathbb{E}_{x \sim p(x)} \sum_{i=1}^k \sum_{m=1}^{k} \Big[ P(y = i\mid Y = m, s = 0,  x)  P (Y=m, s=0 \mid x) \Big] \mathcal{L}[f(x), i] \\
             &= \mathbb{E}_{(x, Y, s=1)} \mathcal{L}[f(x), i] +  \mathbb{E}_{(x, Y, s=0)} \sum_{i=1}^{k} P(y=i \mid Y, s=0, x) \mathcal{L}[f(x), i] \\
 \end{aligned}
 \end{equation}

 \subsection{A.3 The details of datasets}
 \label{app:datasets}
 In this section, we provide a detailed description of datasets used in our experiments.
 \begin{itemize}
     \item CIFAR-100: A coarse-grained dataset comprising 60,000 color images divided into 100 classes, with 600 images per class, 500 training and 100 test, and a resolution of $32 \times 32$ pixels, covering everyday objects and animals.
     \item Tiny-ImageNet: A coarse-grained dataset as a subset of ImageNet, containing 200 classes with 550 images per class, 500 training and 50 validation, and a resolution of $64 \times 64$ pixels, more challenging than CIFAR-100 due to smaller size and more classes.
     \item Caltech-101: A coarse-grained dataset with images from 101 object categories plus one background category, totaling around 9,000 images, 40–800 per class, split into 70\% training and 30\% testing.
     \item Food-101: A fine-grained dataset consisting of 101 food categories with 1,000 images per class, 101,000 total, split into 75,750 training and 25,250 test images, designed for fine-grained food classification.
     \item EuroSAT: A fine-grained satellite image dataset with 27,000 labeled images across 10 land use/cover categories, each $64 \times 64$ and derived from Sentinel-2 data, split into 70\% training and 30\% testing.
     \item Describable Textures Dataset (DTD): A fine-grained dataset comprising 5,640 texture images categorized into 47 classes such as “striped,” “dotted,” and “zigzagged”, supporting fine-grained texture classification, split into 70\% training and 30\% testing.
 \end{itemize}

 \subsection{A.4 The details of prompts}
 \label{app:prompts}
 This section provides detailed descriptions of the prompts used in our experiments.

 Prompt A and Prompt B share the same semantic structure but are written in different languages. The prompt consists of the following instructions:

 \begin{itemize}
     \item Step 1: Carefully analyze the main object and details in the image.
     \item Step 2: Compare the image with the following candidate labels and select the best match.
     \item Step 3: Output only the numeric index of the most suitable label, without any additional text, punctuation, or explanation. The output should be an integer between 0 and \texttt{num-classes}.
 \end{itemize}

 Prompt C is a simplified version written in concise English:

 \begin{quote}
 Classify this image. Choose the best matching label from the complete set of ground-truth for the target dataset and return only the index (0 to \texttt{num-classes}) of your choice. Do not return any text, words, or explanation.
 \end{quote}

\end{document}